\documentclass{article}

\usepackage{microtype}
\usepackage{graphicx}
\usepackage{subfigure}
\usepackage{booktabs} % for professional tables
\usepackage{amsmath} % for equations
\usepackage{array}

\usepackage{hyperref} 

\usepackage[accepted]{icml2019}

\begin{document}

\twocolumn[
\title{Joint Repetition Suppression and Content Moderation of Large Language Models}
\date{\vspace{-0.2in}}
\maketitle

%//////////////////// Author information ////////////////////////

% Include the name, group and email of the author.
% Use the standard \icmlauthor{Name, Group, email@email.com}{}\\
% Dont remove the curly brackets {} at the end.

% Equal contribution.
% Adding "equal" to the second bracket will create an * meaning equal contribution.
% If you are using equal contribution read further instructions aftyer square bracket ].
% Affiliations will be numbered in order of appearance
\icmlsetsymbol{equal}{*}

\begin{icmlauthorlist}
\icmlauthor{Minghui Zhang, Word + Editor, minghuizhang@microsoft.com}{}\\
\icmlauthor{Alex Sokolov, Word + Editor, alexsokolov@microsoft.com}{}\\
\icmlauthor{Weixin Cai, Word + Editor, weixincai@microsoft.com}{}\\
\icmlauthor{Si-Qing Chen, Word + Editor, sqchen@microsoft.com}{}\\
\end{icmlauthorlist}

\vspace{0.4in}
\begin{abstract}
Natural language generation (NLG) is one of the most impactful fields in NLP, and recent years have witnessed its evolution brought about by large language models (LLMs). As the key instrument for writing assistance applications, they are generally prone to replicating or extending offensive content provided in the input. In low-resource data regime, they can also lead to repetitive outputs \cite{holtzman2019curious}. Usually, offensive content and repetitions are mitigated with post-hoc methods, including n-gram level blocklists, top-k and nucleus sampling. In this paper, we apply non-exact repetition suppression using token and sequence level unlikelihood loss, and further explore the framework of unlikelihood training objective in order to jointly endow the model with abilities to avoid generating offensive words and phrases from the beginning. Finally, with comprehensive experiments, we demonstrate that our proposed methods work exceptionally in controlling the repetition and content quality of LLM outputs.
\\

\end{abstract}
\vspace{0.4in}
]

% This command creates the footnote for the equal contribution*
% If you need to mention equal contribution uncoment the next line
%	\printAffiliationsAndNotice{\icmlEqualContribution} % Uncomment for equal contribution
\clearpage

\section{Introduction}
\label{introduction}

Over the years, large language models have become more impactful as they are being applied to an increasing range of features and products \cite{brants2007large, brown2020language}. LLMs have been widely employed in various writing assistance applications, aiding users in generating high-quality, human-like text. However, despite their immense potential, LLMs often struggle with certain limitations that prevent them from further mimicking actual human-written content. Specifically, generations from LLMs sometimes contain repetitions \cite{holtzman2019curious} and controversial phrases (political topics, racial topics, etc.). Repetition refers to the tendency of LLMs to produce sentences or phrases that are either identical or very similar to each other within the generated text. This can lead to a lack of diversity in the output and cause the generated content to appear unnatural, monotonous, or even nonsensical. The problem of repetition can be particularly severe in low-resource data regimes where the model has limited training data to learn from, leading to a higher likelihood of generating repetitive content. Another significant challenge associated with LLMs is the generation of controversial or offensive phrases. LLMs are trained on vast amounts of text data from the internet, and as a result, they may inadvertently learn to generate content that is politically biased, racially insensitive, or otherwise offensive. This can pose a significant risk for applications that rely on LLMs for content generation, as it may lead to the dissemination of harmful or inappropriate content.

Currently, mainstream solutions for both problems are post-hoc, rule-based methods, including n-gram level blocklists, top-k and nucleus sampling \cite{holtzman2019curious}, which are suboptimal. Not only are they not effective enough - words may go through post-filters if they are not exact matches, but they also degrade the output by breaking its coherence.

To overcome these limitations, in this paper we first review the effectiveness of repetition removal method based on sentence embedding, and then we adopt token and sequence level unlikelihood training objective \cite{welleck2019neural} as the non-exact repetition suppression solution. Furthermore, we extend the unlikelihood training objective and customize it to jointly address a broader range of problems. Finally, we demonstrate through experiments that our approach Pareto-dominates the baseline method and can help suppress repetition and moderate content with minimum impact to model performance.

\section{Related Work}
\label{related_work}

Various techniques have been proposed to tackle the repetition problem in language models. Some of these methods include modifying the decoding algorithm, such as using nucleus sampling \cite{holtzman2019curious}, top-k sampling \cite{fan2018hierarchical} and diverse beam search \cite{vijayakumar2016diverse}. Another approach to suppress repetition is to penalize repeated n-grams during the decoding process. Holtzman et al. \cite{holtzman2019curious} acknowledge the problem of repetition and propose a new decoding strategy to mitigate it. Welleck et al. \cite{welleck2019neural} try to tackle this problem by customizing training objectives. Fu et al. \cite{fu2021theoretical} provide a theoretical framework to systematically analyze the root cause of repetition and the reason behind existing techniques.

Content moderation in LLMs aims to prevent the generation of offensive or controversial content. Several methods have been proposed to mitigate this issue, including the use of blocklists, adversarial training, reinforcement learning from human feedback \cite{ziegler2019fine}, or rule-based filters. In practice, post-processing filters like blocklists and offensiveness classifiers are still most economic and prevalent.

\section{Non-Exact Repetition Suppression}
\label{non-exact_repetition_suppression}

Since its inception, BERT \cite{devlin2018bert} has been used as a tool for encoding sentences and generating their representations. However, BERT isn’t pre-trained to be natively support sentence similarity detection. SentenceBERT \cite{reimers2019sentence}, being a solution to this issue, makes structural changes on top of BERT for training and inference respectively with cosine-similarity in mind. In practice, we calculate the similarity score for any two sentences in a paragraph generated by the LLM, and only keep one sentence in any pair with a score above a pre-determined threshold. This SentenceBERT-based method is a great representative of repetition detection \& removal approaches. Figure 1 contains the architectures of SentenceBERT during training (left) and inference (right).

\begin{figure}[ht]
\begin{center}
\centerline{\includegraphics[width=\columnwidth]{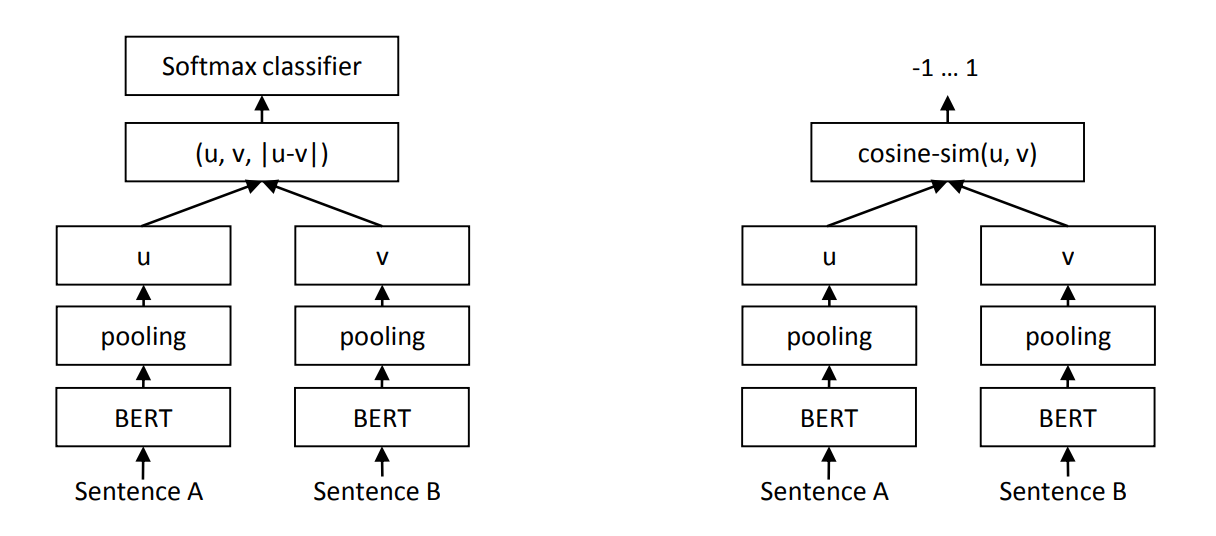}}
\caption{Architecture of SentenceBERT\cite{reimers2019sentence}}
\label{sbert_arch}
\end{center}
\vskip -0.3in
\end{figure}

As we discussed above in the introduction, repetition prevention is a more ideal way to address the issue than post-hoc repetition removal. We adopt the unlikelihood (UL) training objective in Welleck et al. \cite{welleck2019neural}. Specifically, ordinary likelihood training objective minimizes the loss for ground truth tokens:
\begin{equation}
    \mathcal{L}_{Likelihood}^{t}(p_\theta(\cdot|x_{<t}))=-\log p_\theta(x_t|x_{<t})
\end{equation}

Where $p_\theta$ denotes the language model with parameters $\theta$, and $t$ is the current time step. 

In unlikelihood training objective, higher loss is applied to a set of tokens, called negative candidates, and thus discourages the model from generating them. Token-level UL objective defines negative candidates $C^t_{UL-token}$ as previous tokens within the target (ground truth).
\begin{multline}
    \mathcal{L}_{Unlikelihood}^{t}(p_\theta(\cdot|x_{<t}), C^t)= \\
        -\alpha\sum_{c\in C^t} \log ( 1-p_\theta(c|x_{<t}) ) - \log p_\theta(x_t|x_{<t}) \label{eq:token_ul}
\end{multline}
\begin{equation}
    C^t = C^t_{UL-token} = \{x_1,\ldots,x_{t-1}\}\setminus\{x_t\}
\end{equation}

Coefficient $\alpha$ controls the proportion of the unlikelihood loss term.

Sequence-level UL objective follows the same principle, except that it defines negative candidates as previous duplicate n-grams within the predicted sequence.
\begin{equation}
    \begin{gathered}
        C^t = C^t_{UL-seq} = \{x_t\} \text{ if } (x_{t-i},\ldots,x_{t+j})\in x_{<t-i} \\
         \text{ for any } i+j+1 = n
    \end{gathered}
\end{equation}

\section{Content Moderation}
\label{content_moderation}

The unlikelihood training objectives, with both token-level UL loss and sequence-level UL loss, are not standalone loss functions. In fact, they propose a platform with the concept of negative candidates, and having repetition-related context can be seen as a special case. Based on this idea, we further generalize the scope of negative candidates and define a new loss term called “block loss”.  In block loss, the negative candidates are defined as blocklist phrases $B_n$ that we have collected in advance.
\begin{equation}
    C^t = C^t_{block} = \bigcup^{10}_{n=2}C^t_{block-n}
\end{equation}
\begin{equation}
    \begin{gathered}
        C^t_{block-n} = \{x_t\} \text{ if } (x_{t-i},\ldots,x_{t+j})\in B_n \\
         \text{ for any } i+j+1 = n
    \end{gathered}
\end{equation}

\section{Experiments}

First, we evaluate the performance of SentenceBERT-based post-processing methods. Then, we compare different aspects of unlikelihood training objectives. Lastly, we show the efficacy of our novel block loss in the setting of sequence-level UL training objective. 

\begin{table*}[t]
\vskip 0.1in
\begin{center}
\begin{small}
\begin{tabular}{lcccc}
\toprule
Model & S-BERT threshold & Repeated Word Ratio & Repeated 4-word Phrase Ratio & BLEURT \\
\midrule
Baseline            & -    & 40.553 & 0.017 & 0.501 \\
Baseline + S-BERT   & 0.91 & 40.419 & 0.017 & 0.501 \\
Baseline + S-BERT   & 0.8  & 39.171 & 0.013 & 0.500 \\
Token-level UL Obj  & -    & 34.463 & 0.011 & 0.512 \\
\bottomrule
\end{tabular}
\end{small}
\end{center}
\caption{Results for SentenceBERT and token-level unlikelihood objective}
\vskip -0.1in
\label{table:sbert}
\end{table*}

\begin{table*}[t]
\vskip 0.1in
\begin{small}
\begin{tabular}{lccccccccm{0.07\textwidth}<{\centering}m{0.07\textwidth}<{\centering}}
\toprule
Model & ppl & acc & rep-128 & wrep-128 & seq-rep-1 & seq-rep-4 & uniq & uniq-seq & human-seq-rep-1 & human-seq-rep-4 \\
\midrule
Baseline           & 11.786 & 0.521 & 0.795 & 0.586 & 0.667 & 0.506 & 7332 & 6929 \\
Token-level UL Obj & 10.201 & 0.533 & 0.390 & 0.182 & 0.585 & 0.394 & 7468 & 7035 & 0.317 & 0.022 \\
Seq-level UL Obj   & 11.382 & 0.524 & 0.366 & 0.168 & 0.311 & 0.059  & 7491 & 7152 \\
\bottomrule
\end{tabular}

\begin{tabular}{lcccccm{0.102\textwidth}<{\centering}m{0.102\textwidth}<{\centering}}

Model & ROUGE1-F & ROUGE2-F & ROUGEL-F & BERTScore & BLEURT & perc-grammar-wrong & count-grammar-wrong\\
\midrule
Baseline           & 0.496 & 0.207 & 0.324 & 0.289 & 0.501 & 0.034 & 0.145 \\
Token-level UL Obj & 0.497 & 0.212 & 0.332 & 0.310 & 0.512 & 0.021 & 0.087 \\
Seq-level UL Obj   & 0.486 & 0.212 & 0.340 & 0.325 & 0.509 & 0.023 & 0.093 \\
\bottomrule
\end{tabular}
\end{small}
\caption{Results for unlikelihood objectives}
\vskip -0.1in
\label{table:ul}
\end{table*}

\subsection{Experimental Setup}

We use a sequence-to-sequence text rewriting task to evaluate the approaches discussed in the sections above, with bullet points as inputs and paragraphs as outputs. In practice, this task could translate to writing assistant products or features capable of generating fluent documents based on user-provided outlines.

We use pre-trained BART model \cite{lewis2019bart} as the common starting point and apply different training objectives during fine-tuning. The baseline model is fine-tuned until convergence with regular likelihood loss. Another model is fine-tuned from the starting point with token-level UL objective. The last model is fine-tuned with sequence-level UL objective on top of the previous model. Following Welleck et al. \cite{welleck2019neural}, sequence-level UL objective is applied 50\% of the time during fine-tuning, with the other 50\% of the time using model's original token-level training objective. 

Inputs are encoded by SentencePiece tokenizer \cite{kudo2018sentencepiece}. Teacher forcing \cite{williams1989learning} is applied during training to ensure a faster and more stable convergence. We use beam search as the decoding strategy with beam size 5 across all experiments. All training is done on an Azure ML Compute instance with 4 Nvidia V100 16GB GPUs.

\subsection{Data}

Our data is in a simple format, as each data sample consists of one paragraph as ground truth and its corresponding list of bullet points in pure text format. The bullet is denoted by an asterisk ($*$), and the sub-bullet is denoted by double asterisk ($**$). The size of the training set, validation set, and test set are 45k, 6.6k, 1.4k respectively. Bullet points are summarized by human annotators from paragraphs in public Word documents corpus, and unqualified samples (too long, offensive, etc.) are removed in advance. Below is one example sample of the dataset:
{\footnotesize
\begin{verbatim}
{"bullet_points": "Evans, CMO of Subaru of 
America, Inc. (SOA) * 20+yrs experience * 
oversaw demand generation, brand awareness 
* customer engagement programs * SOA 
revenue increase ** 40% to 65%of parent 
company", 

"paragraph": "Evans boasts over 20 years of 
experience in the automotive industry, most 
recently as CMO of Subaru of America, 
Inc. (SOA). In that position, he oversaw 
demand generation, brand awareness and 
customer engagement programs that directly 
contributed to the company's highest years 
of sales growth in its history. SOA's 
revenue increased from 40% to 65% of parent 
company Fuji Heavy Industries' (FHI) 
overall revenue during Evans' tenure, with 
FHI's stock the best performer on the 
Nikkei."}
\end{verbatim}}
For the evaluation of sequence-level training objective with block loss, we use a subset of the original dataset, with 50\% samples containing blocklist phrases \& 50\% regular samples, in order to better demonstrate the results and save computational cost.

\subsection{Evaluation Metrics}

We record an extensive range of metrics during evaluation in four categories: language model metrics, repetition metrics, summarization/translation-based metrics, and grammatical metrics. Below is a list of itemized metric descriptions.

\textbf{Language model metrics} measure the basic capabilities of language models, namely how well they can predict the next token.
\begin{itemize}
  \item Perplexity (ppl) 
  \item Next-token prediction accuracy (acc) 
\end{itemize}

\textbf{Repetition metrics} measure the duplicativeness and uniqueness of generation on both token level and sequence level.
\begin{itemize}
  \item Repetition (rep-$l$): the fraction of next-token prediction that occur in previous $l$ tokens 
  \item Wrong repetition (wrep-$l$): similar to rep-$l$, but only counts token repeats that are not equal to gold next token 
  \item Portion of duplicate 1-grams (seq-rep-1) 
  \item Portion of duplicate 4-grams (seq-rep-4) 
  \item Number of unique next-token predictions (uniq) 
  \item Number of unique tokens in the generated paragraph (uniq-seq)
\end{itemize}

\textbf{Summarization/translation-based metrics} measure the semantic consistency between model's input and output.
\begin{itemize}
  \item ROUGE \cite{lin2004rouge}: similarity between prediction \& ground truth measured with overlap n-grams \& longest common subsequence 
  \item BERTScore \cite{zhang2019bertscore}: similarity between prediction \& ground truth measured with BERT representations 
  \item BLEURT \cite{sellam2020bleurt}: a transfer learning-based NLG metric 
\end{itemize}

\textbf{Grammatical metrics} measure the prevalence of grammar mistakes in model outputs, and thus monitor the impact on grammatical performance.
\begin{itemize}
  \item Percentage of sentences with grammar mistakes (perc-grammar-wrong) 
  \item Number of grammar mistakes per sentence (count-grammar-wrong)
\end{itemize}

\subsection{Results}

In table \ref{table:sbert}, SentenceBERT, as a post-hoc method, is Pareto dominated by unlikelihood objectives regardless of former's threshold choice, which justifies our transition from post processing to custom training objective. In table \ref{table:ul}, token-level UL objective significantly reduces token repetition, while additional training with sequence-level UL objective further reduces repetitive n-grams by a huge margin. Overall, unlikelihood training objectives help suppress repetition with minimum impact to performance. 

Table \ref{table:examples_ul} lists a set of generations from different models given the same outline. Baseline output suffers from severe repetition problem whereas the outputs of models trained with unlikelihood objectives do not, and thus provide a better text semantically and syntactically. Outline contains additional line breaks and indentations for better illustration. Ground truth (the original paragraph) is also attached for reference. 

In terms of training time, as shown in table \ref{table:ul_time}, token-level unlikelihood objective has a similar result as regular likelihood objective, which is expected since the former can reduce to the latter when $\alpha = 0$ in equation (\ref{eq:token_ul}). In contrast, sequence-level UL objectives are much more time-consuming. This order-of-magnitude difference between two levels can be explained by the iterative n-gram checks during the identification of sequence-level negative candidates. 

In table \ref{table:block}, sequence-level UL objective with our novel block loss reduces the number of outputs containing unwanted blocklist phrases. Similarly, a set of generations from different models as well as corresponding ground truth are listed in table \ref{table:examples_block}. The word "abortion" is in the blocklist and successfully avoided in the output of the model fine-tuned with sequence-level UL objective with block loss.

\begin{table}[t]
\vskip 0.1in
\begin{center}
\begin{small}
\begin{tabular}{lc}
\toprule
Model & Training Time \\
\midrule
Baseline                           & 25 \\
Token-level UL Obj                 & 25 \\
Seq-level UL Obj                   & 729 \\
Seq-level UL Obj with block loss   & 530 \\
\bottomrule
\end{tabular}
\end{small}
\end{center}
\caption{Training time (seconds/100 steps) of unlikelihood objectives}
\vskip -0.1in
\label{table:ul_time}
\end{table}

\begin{table}[t]
\vskip 0.1in
\begin{center}
\begin{small}
\begin{tabular}{lcm{0.13\textwidth}<{\centering}cccccccc}
\toprule
Model & beta & \# of outputs with blocklist phrases \\
\midrule
Baseline                         & -    & 119 \\
Seq-level UL Obj with block loss & 10   & 84  \\
\bottomrule
\end{tabular}
\end{small}
\end{center}
\caption{Results for sequence-level unlikelihood objective with block loss}
\vskip -0.1in
\label{table:block}
\end{table}

\begin{table*}[t]
\begin{center}
\begin{small}
\begin{tabular}{m{0.2\textwidth}m{0.7\textwidth}}
\toprule
Outline & PAML\newline* healthcare solutions company focused on diagnostics\newline\hspace*{2em}** recognized for innovations\newline* joint venture partnerships leader w/ community hospitals \\
\midrule
Baseline & PAML is a healthcare solutions company focused on diagnostics and has been recognized for its innovations in \underline{\smash{diagnostic and diagnostic}} technology. PAML has been a joint venture partnerships leader with community hospitals in the United States, Canada, Australia, and the United Kingdom. In addition to providing a wide range of diagnostic services to patients and their families, PAML has also been recognized as one of the world's leading healthcare solutions companies in the field of \underline{\smash{diagnostic, diagnostic, and diagnostical diagnostic}} technologies. \\
\midrule
Token-level UL Obj & PAML is a healthcare solutions company focused on diagnostics. PAML has been recognized as one of the world's leading innovators in the field of medical imaging and medical diagnostic technology. The company's Joint Venture Partnerships are a leader in partnerships with community hospitals across the United States and internationally. The joint venture partnerships provide a unique opportunity to collaborate with a wide range of healthcare providers across the globe. \\
\midrule
Seq-level UL Obj & PAML is a healthcare solutions company focused on diagnostics. PAML has been recognized as one of the world's leading innovators in the field of diagnostic technology. The company's Joint Venture Partnerships is a leader in partnerships with community hospitals across the U.S. and internationally. The joint venture partnerships are designed to provide high-quality, high-performing diagnostic services to patients with complex medical conditions.\\
\midrule
Ground Truth & PAML, LLC, a healthcare solutions company with a focus in diagnostics, is headquartered in Spokane, WA. One of the top reference laboratories in the country, PAML is also nationally recognized for innovations in clinical diagnostics. PAML is also an industry leader in joint venture partnerships with community-based hospitals because of their expertise in developing cutting-edge laboratory management systems, including state-of-the-art automation and support services. PAML is a founding member of PACLAB, with the largest outreach lab market share in the region. \\
\bottomrule
\end{tabular}
\end{small}
\end{center}
\caption{Example outputs from models fine-tuned with unlikelihood objectives}
\label{table:examples_ul}
\end{table*}

\begin{table*}[t]
\begin{center}
\begin{small}
\begin{tabular}{m{0.2\textwidth}m{0.7\textwidth}}
\toprule
Outline & Abortion\newline * Malta = illegal\newline\hspace*{2em}** explains increase in deaths due congenital malformations\newline * where legal\newline\hspace*{2em}** pregnancies are terminated b4 fetus is delivered \\
\midrule
Baseline & In Malta, \underline{\smash{abortion}} is illegal. This explains the increase in the number of deaths due to congenital malformations. Where legal, pregnancies are terminated b4 the fetus is delivered. In the case of a woman who is pregnant at the time of the abortion, the pregnancy is terminated, and the mother is not able to give birth to the child. The mother is unable to provide the child with a safe, safe, and safe environment for the delivery of the child.\\
\midrule
Seq-level UL Obj with\newline Block Loss & In the Republic of Malta, it is illegal to carry out an unplanned pregnancy without the consent of the mother. This explains why there has been an increase in deaths due to congenital malformations. However, in other countries where it is legal, pregnancies are terminated after the fetus is delivered. This is not the case in England and Wales, where it has been shown that there are many cases where women who do not have access to medical care are forced to leave the country.\\
\midrule
Ground Truth & The comparability of these figures to those of other countries is undermined by the fact that, in contrast to many other countries, abortion in Malta is illegal. This may partly explain the increase in the proportion of deaths due to congenital malformations, deformations and chromosomal abnormalities (Q00-Q99), which are largely unpreventable. Whereas in countries where abortion is legal these pregnancies are likely to be terminated before the fetus is delivered, in Malta such cases may possibly live for a few minutes, hours or days, after delivery, in which case they are included in the perinatal, neonatal, infant and under-five mortality rates.\\
\bottomrule
\end{tabular}
\end{small}
\end{center}
\caption{Example outputs from model fine-tuned with sequence-level unlikelihood objective with block loss}
\label{table:examples_block}
\end{table*}

\section{Conclusions \& Future Work}

In this paper, we explored a joint effort of non-exact repetition suppression and content moderation to address the limitations of LLMs in generating repetitive and offensive content. We first set up SentenceBERT as a baseline of repetition detection \& post-processing methods. Then, we utilized multiple levels of unlikelihood training objectives to suppress repetition at the step of generation. Finally, we exploited and further generalized the unlikelihood training objective and brought it into the field of content moderation. We demonstrated that our proposed methods work exceptionally well in controlling the repetition and content quality of LLM outputs. The results showed that multi-level unlikelihood training objectives and our novel block loss greatly reduce token repetition, repetitive n-grams and offensive content while maintaining minimum impact on model performance.

In the future, we plan to explore more possibilities for the unlikelihood objective framework and negative candidates, as well as expanding the potential applications of our proposed methods in more complex NLP scenarios. Additional investigation could be performed on the impact of the size and prevalence of the blocklist.

\clearpage
\clearpage
\bibliographystyle{ieeetr}
\bibliography{paper}

\end{document}